\documentclass[10pt,twocolumn,letterpaper]{article}

\usepackage[pagenumbers]{cvpr} 
\usepackage{amsmath, amssymb}

\usepackage{amsmath}
\DeclareMathOperator*{\argmax}{argmax} 

\definecolor{cvprblue}{rgb}{0.21,0.49,0.74}
\usepackage[pagebackref,breaklinks,colorlinks,allcolors=cvprblue]{hyperref}

\def\paperID{13098} 
\def\confName{CVPR}
\def\confYear{2026}

\title{UTrice: Unifying Primitives in Differentiable Ray Tracing and Rasterization via Triangles for Particle-Based 3D Scenes}

\author{Changhe Liu \quad Ehsan Javanmardi \quad Naren Bao \quad Alex Orsholits \quad Manabu Tsukada\footnotemark[1]\\
The University of Tokyo\\
{\tt\small \{lch01234, ejavanmardi, naren, alex-orsholits, mtsukada\}@g.ecc.u-tokyo.ac.jp}
}
\begin{document}
\twocolumn[{%
\renewcommand\twocolumn[1][]{#1}%
\maketitle
\vspace{-2em}
\includegraphics[width=\linewidth,trim=0 80 0 120,clip]{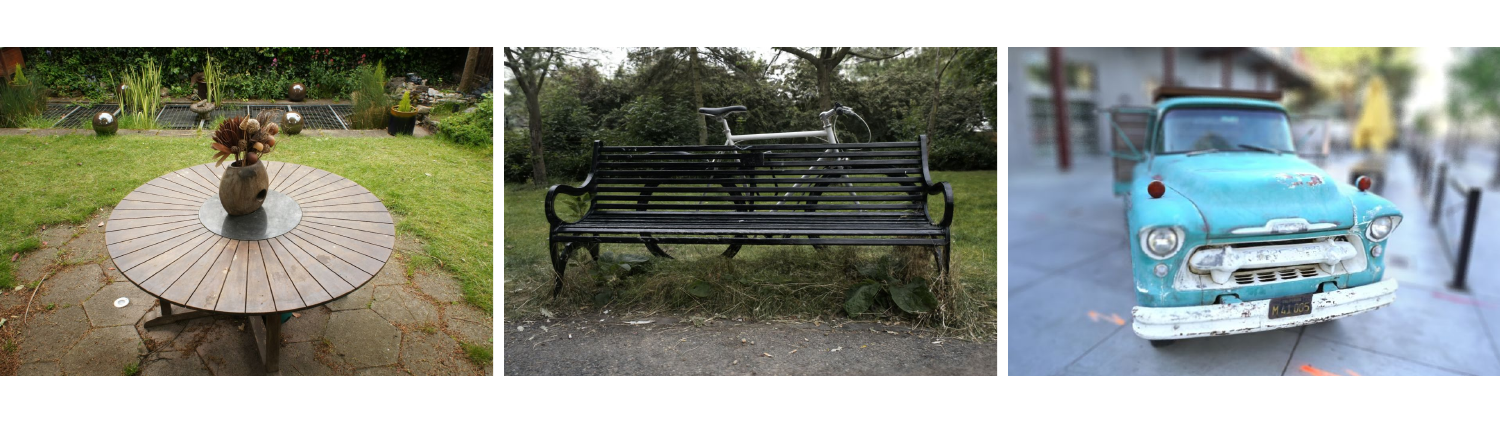}
\captionof{figure}{{\bf Rendering results of our method.}
By directly using triangles as primitives, UTrice achieves higher rendering quality than 3DGRT~\cite{3dgrt} while maintaining real-time performance (left and middle), and can produce realistic ray tracing effects such as depth of field (right). It also renders triangles optimized by 3DTS~\cite{3dts} without additional processing, unifying the primitives used in rasterization and ray tracing based optimization.}
\vspace{2em}
\label{fig:teaser}
}]

\iftoggle{cvprfinal}{
\renewcommand{\thefootnote}{\fnsymbol{footnote}}
\footnotetext[1]{Corresponding author}
}

\begin{abstract}

Ray tracing 3D Gaussian particles enables realistic effects such as depth of field, refractions, and flexible camera modeling for novel-view synthesis. However, existing methods trace Gaussians through proxy geometry, which requires constructing complex intermediate meshes and performing costly intersection tests. This limitation arises because Gaussian-based particles are not well suited as unified primitives for both ray tracing and rasterization.
In this work, we propose a differentiable triangle-based ray tracing pipeline that directly treats triangles as rendering primitives without relying on any proxy geometry. Our results show that the proposed method achieves significantly higher rendering quality than existing ray tracing approaches while maintaining real-time rendering performance. Moreover, our pipeline can directly render triangles optimized by the rasterization-based method Triangle Splatting, thus unifying the primitives used in novel-view synthesis.\iftoggle{cvprfinal}{
    Code is available at %
    \href{https://github.com/waizui/UTrice}{\textcolor{blue}{https://github.com/waizui/UTrice}}.
}

\end{abstract}
\section{Introduction}
\label{sec:intro}
Novel-view synthesis has advanced rapidly in recent years, and 3D Gaussian Splatting~\cite{Kerbl2023-nj} has attracted significant attention for its outstanding rendering quality and real-time performance. Compared with the original work, recent 3D Gaussian Splatting methods have made significant progress in both rendering efficiency, memory usage and fidelity. Moreover, a growing number of studies have focused on developing more effective and generalizable particle representations. 2DGS~\cite{2dgs} replaced the 3D Gaussians with 2D oriented planar Gaussian disks, and further introduced a distortion regularization term. Owing to the better view-consistent geometry of 2D Gaussians, their method achieved the highest reconstruction accuracy. Triangle Splatting(3DTS)~\cite{3dts}, introduced the use of triangles as primitives for optimization. Inspired by 2D Gaussian Splatting, they incorporated a distortion regularization term and replaced the heuristic densification strategy with an MCMC-based densification~\cite{3dgsmcmc} approach. Their method achieved superior performance in terms of fidelity, training speed, and rendering throughput. More importantly, as triangles are the most universal primitives in computer graphics, this representation can be easily integrated into existing rendering applications.

On the other hand, ray tracing is a widely used rendering technique in computer graphics. It is employed to perform Monte Carlo integration of the Light Transport Equation~\cite{Kajiya86}, providing highly realistic, physically based rendering results. 3DGRT~\cite{3dgrt} integrates ray tracing into 3D Gaussian Splatting. Their pipeline allows for efficient utilization of hardware-accelerated ray tracing~\cite{OptiX}. They enclose each Gaussian particle with proxy geometry (an icosahedron) and use a buffer to store up to 
$k$ ray–particle intersections in front-to-back order during BVH traversal, enabling efficient data access. Due to this front-to-back traversal design, they re-derived the gradients of the optimizable parameters.

However, 3DGRT does not trace primitives directly, leading to additional overhead in intersection testing and BVH construction, as well as extra memory usage for proxy geometry. These costs are significant; for example, BVH building can occupy a large portion of the total runtime~\cite{3dgrt}.
Inspired by Triangle splatting \cite{3dts}, our work replaces the ray tracing primitives with differentiable triangles, eliminating the need for proxy geometry. This design substantially reduces the computational cost of BVH construction and intersection testing, enabling a framework capable of more realistic and generalizable rendering, as illustrated in Figure~\ref{fig:teaser}.

In summary, our main contributions are as follows:
\begin{itemize}
    \item We propose an optimization pipeline for ray-traced, particle-based radiance fields that uses triangles as primitives.
    \item We introduce a differentiable formulation that enables triangles to serve as optimizable primitives within the ray tracing pipeline.
    \item We demonstrate that using triangles as primitives yields better generality and higher rendering quality compared to Gaussian-based representations.
\end{itemize}


\section{Related Works}
\label{sec:relatedwork}
{\bf Novel view synthesis.} Neural Radiance Fields (NeRF)~\cite{nerf} utilize an implicit multi-layer perceptron (MLP) that takes 3D positions and viewing directions as inputs to predict density and view-dependent color for differentiable volume rendering, revolutionizing the field of novel view synthesis. The high fidelity of NeRFs has made them the de facto representation for novel view synthesis. Subsequent works~\cite{mipnerf,mignerf360,zipnerf} focus on further improving rendering quality. However, despite their photo-realistic quality, the training and rendering performance of NeRF remain insufficient for real-time applications. Subsequent works have therefore focused primarily on addressing these limitations ~\cite{instant-neural,kiloNeRF,tensorRF,directVoxelgrid},~ \cite{mobnerf} introduced a polygon-based representation of NeRFs, where a small MLP is executed within a GLSL fragment shader to predict colors, achieving over a 10$\times$ rendering speed improvement compared to their baseline on mobile devices.

\noindent{\bf Point-based differentiable rendering.}
3D Gaussian Splatting~\cite{Kerbl2023-nj} represents scenes with anisotropic Gaussian particles and uses a tile-based rasterizer to optimize their position and appearance, achieving state-of-the-art novel-view synthesis and real-time rendering. This work brings differentiable rendering to real-world applications. Subsequent works~\cite{scaffoldgs,taming3dgs,3dgsmcmc} have focused on developing optimized training strategies to improve the memory efficiency and rendering quality of 3DGS. Later, the works of~\cite{4dgs,deformgs} achieved the reconstruction of dynamic scenes. Some recent works have experimented with alternative kernels of different characteristics. 2D Gaussian Splatting~\cite{2dgs} represents scenes using planar 2D Gaussian primitives and introduces both distortion loss and normal loss, achieving excellent reconstruction accuracy, BillBoard Splatting~\cite{bbsplat} extends their approach by using learnable textures instead of Gaussians, achieving higher visual realism with fewer primitives.
Triangle Splatting~\cite{3dts} replaces Gaussians with differentiable triangles, better preserving high-frequency details and edge sharpness.~\cite{2dts} uses the Generalized Exponential Function~\cite{ges} defined on 2D triangles as the primitive, achieving good performance in mesh reconstruction. 

\noindent{\bf Differentiable ray tracing of particles.} Some visual effects cannot be achieved merely by changing the kernel function, but rather require specially designed rendering methods — for example, generating effects such as depth of field, environment lighting, or simulating non-pinhole camera models.~\cite{prtgs,tsinghuaprtgs} compute environmental lighting using Precomputed Radiance Transfer (PRT) by introducing additional learnable parameters. \cite{gsshader,gsir,reflectivegs} introduce new material parameters to enable re-shading of 3DGS. 3DGRT~\cite{3dgrt} introduced a ray tracing approach for Gaussian particles, providing a unified pipeline for ray tracing–based effects. Following this,~\cite{envgs} and~\cite{IRGS} adopted their method to enhance the quality of indirect radiance in rendered scenes.~\cite{3dgut} and~\cite{lidarrt} further extended this method to handle non-pinhole camera models, enabling ray tracing–based rendering for panoramic and fisheye imaging systems. 

Existing differentiable ray tracing methods require proxy geometry for each particle, incurring extra computational overhead and complicating integration with rendering pipelines. 
In contrast, we directly use triangles as differentiable primitives, making our approach compatible with modern ray tracing frameworks~\cite{OptiX,dxr} and triangle-based rasterization~\cite{3dts}. 
Removing auxiliary structures eliminates redundant computation and yields higher novel-view synthesis quality.

\begin{figure*}
    \centering
    \includegraphics[width=\linewidth,trim=8 110 8 60,clip]{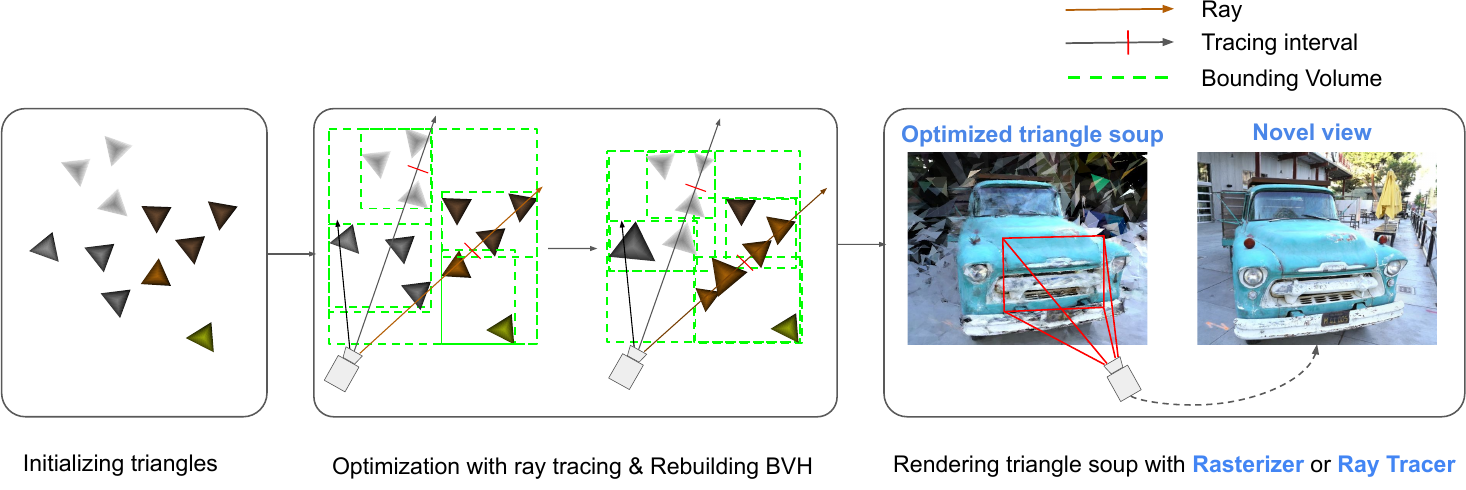}
    \caption{{\bf Overview of the pipeline.} In the leftmost part, the triangles are initialized from the SfM point cloud, where each is generated by sampling three vertices within a unit sphere. They are then passed to OptiX for BVH construction and ray tracing, where each ray is visualized using its corresponding ground-truth pixel color. The triangles are iteratively optimized until the desired rendering quality is achieved and can subsequently be rendered using either a rasterizer or a ray tracer.
}
    \label{fig:pipeline}
\end{figure*}
\section{Method}
\label{sec:method}

\subsection{Background}
{\bf 3D Gaussian splatting.} 3D Gaussian Splatting represents each primitive as an anisotropic Gaussian kernel, defined as
\[
G(x) = \exp\!\left(-\frac{1}{2}(x - \mu)^{\top}\Sigma^{-1}(x - \mu)\right),
\]
where $x \in \mathbb{R}^3$, $\mu \in \mathbb{R}^3$ denotes the particle’s position, and $\Sigma \in \mathbb{R}^{3 \times 3}$ is the covariance matrix. 
During rendering, these particles are first projected into the image space. 
Following~\cite{ewa}, the covariance matrix $\Sigma'$ in camera coordinates is given by
\[
\Sigma' = J\,W\,\Sigma\,W^{\top} J^{\top},
\]
where $J$ is the Jacobian of the affine approximation of the projective transformation, 
and $W$ is the view transformation matrix. 
The overlapping N particles are then composited using the following alpha blending equation to obtain the final pixel color.
\[
\mathcal{C} = \sum_{i=1}^{N} T_i \alpha_i c_i, \qquad 
T_i = \prod_{j=1}^{i-1} (1 - \alpha_j).
\]
where $c_i$ is the color of the $i^{th}$ Gaussian and $T_i$ is the transmittance.

\begin{figure*}
    \centering
    \begin{minipage}[t]{0.5\textwidth}
        \centering
        \includegraphics[width=\textwidth,trim=120 50 120 30,clip]{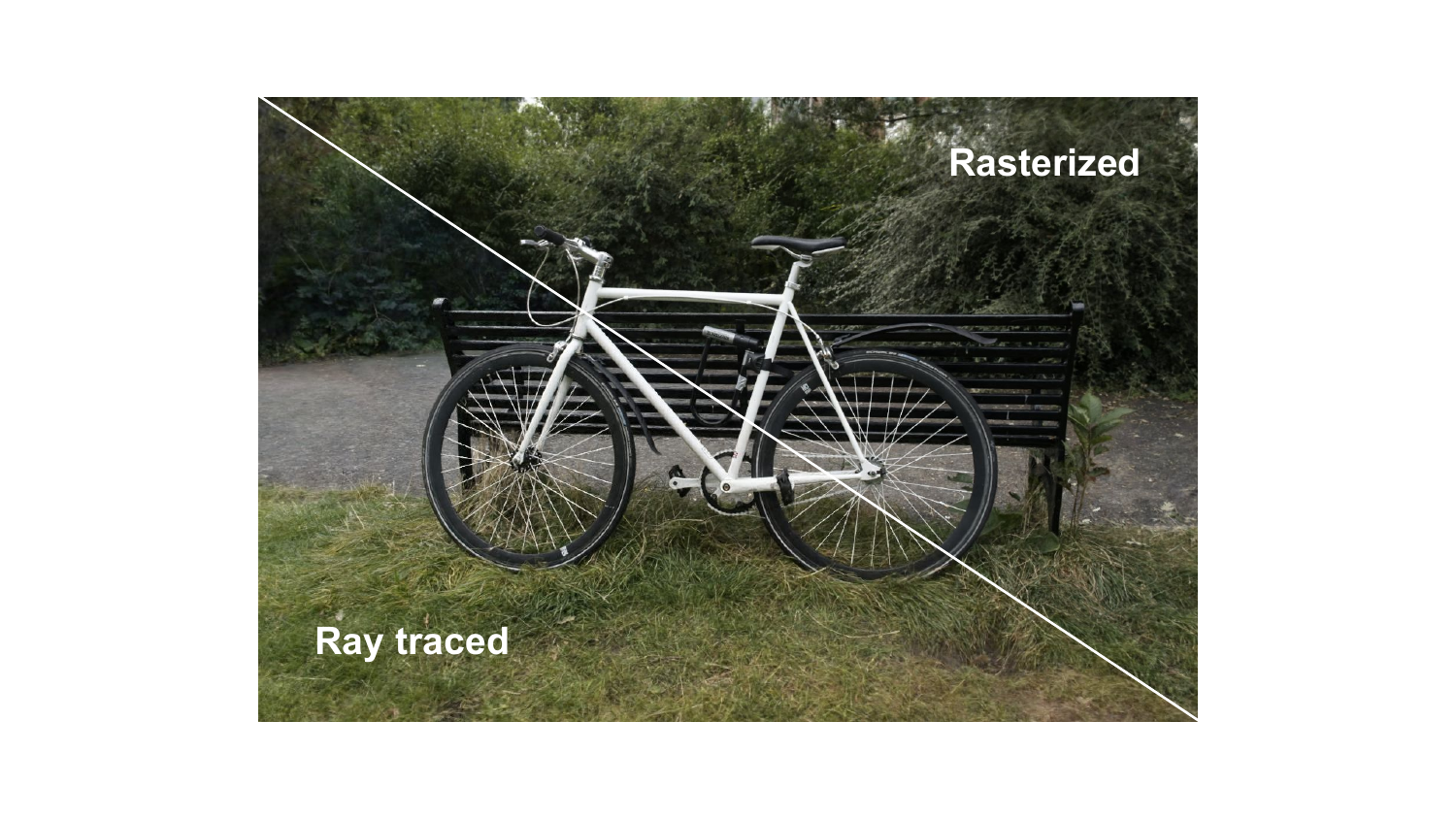}
    \end{minipage}\hfill
    \begin{minipage}[t]{0.5\textwidth}
        \centering
        \includegraphics[width=\textwidth,trim=120 50 120 30,clip]{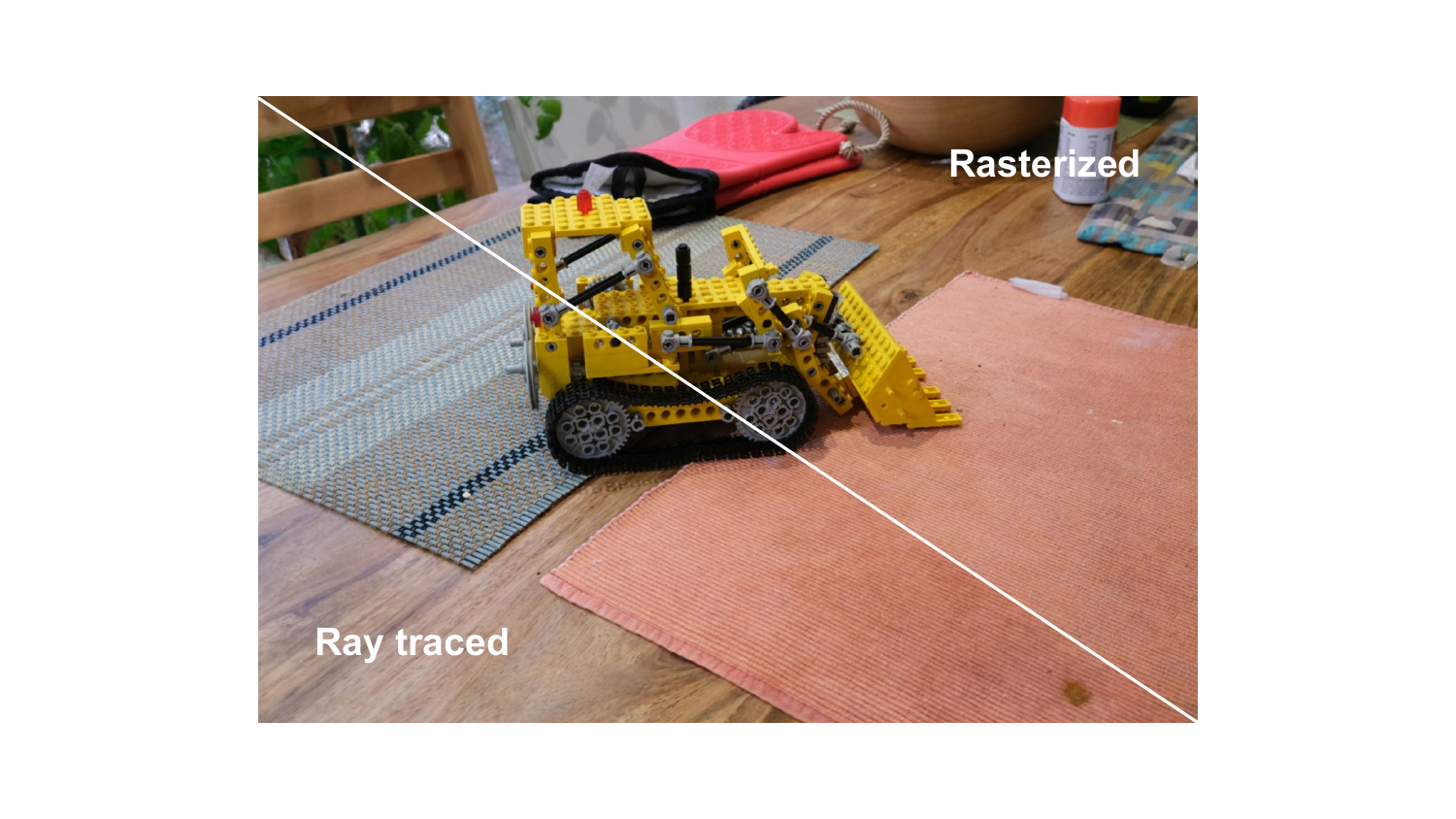}
    \end{minipage}
    \caption{{\bf Compatibility of our ray tracer.} The response of the window function $I$ with the same smoothness factor $\sigma$ remains invariant under linear transformations. Therefore, our tracer can render triangles optimized by rasterization without any additional handling.}
    \label{fig:compatibility}
\end{figure*}

\noindent{\bf Ray tracing Gaussian particles.} In ray tracing, the most fundamental acceleration data structure is the Bounding Volume Hierarchy (BVH). Rays traverse the bounding volumes within the BVH to test for intersections with the contained primitives. 
Under this setting, since a Gaussian kernel is defined over an infinitely smooth convex support, 3DGRT~\cite{3dgrt} uses icosahedra as proxy geometries to tightly enclose each Gaussian particle when building BVH. Following the previous work~\cite{Knoll2019,multialpha}, it employs a k-closest semi-transparent particle-tracing algorithm to retrieve particles that are ordered along the ray direction.

\subsection{Overview of Our Work}
We initialize the triangles from the point clouds obtained via Structure-from-Motion (SfM) following the method of Triangle Splatting~\cite{3dts}. 
We then compute the index buffer for these triangles and use the vertex array together with the index buffer to directly build the BVH in OptiX~\cite{OptiX}. 
During ray tracing, a $k$-element buffer is used to record intersection information, which is iteratively processed until the tracing termination condition is met. 
Afterward, we compute the loss between the rendered images and ground truth using PyTorch~\cite{torch}, and perform backpropagation. 
We implement custom CUDA kernels to compute the gradients of the triangle parameters, and finally optimize all triangle parameters using the Adam optimizer, as illustrated in Figure~\ref{fig:pipeline}.

\subsection{Differentiable Triangle}
\label{sec:differentiable_triangle}
In our work, we choose triangles as the primitives because they are general and simple. 
A triangle is defined by its vertices $v_i \in \mathbb{R}^3,\, i \in \{1,2,3\}$, 
color $c \in \mathbb{R}^3$, smoothness factor $\sigma$, and opacity $o$. 
In implementation, we encode the color using spherical harmonics of degree $3$. 
During rendering, the color of a single pixel is obtained by front-to-back blending of 
all triangles intersected by the ray passing through that pixel. 

\begin{figure}[b]
    \centering
    \includegraphics[width=\linewidth]{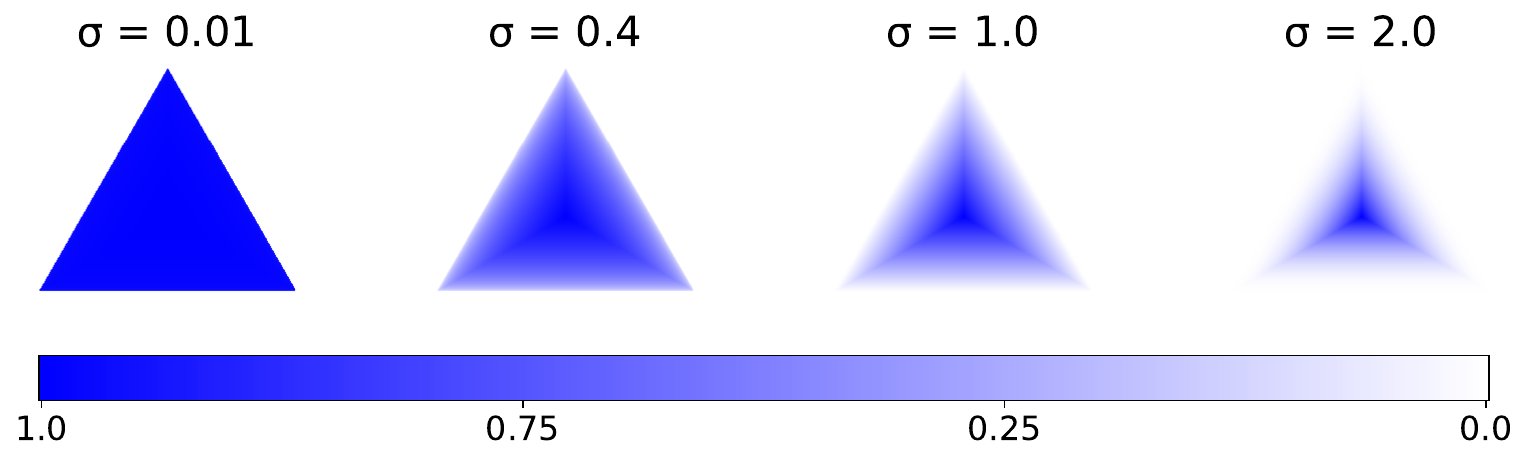}
    \caption{\textbf{Effect of $\sigma$ on the Triangle window function.} Small $\sigma$ (e.g., $0.01$) makes the triangle nearly fully opaque. Larger $\sigma$ increases the sensitivity of the window function to interior-point variations, enabling gradient-based optimization of triangle geometry.}
    \label{fig:windowfunction}
\end{figure}

\noindent{\bf Triangle window function.} We compute the response of a triangle to the ray defined by $r_o + t r_d$ at the intersection point $\mathbf{p}$ as $I(\mathbf{p}) $, where $r_o \in \mathbb{R}^3$ and $r_d \in \mathbb{R}^3$ denote the ray origin and direction, respectively, and $t \in \mathbb{R}$. The function $I$ is a \textit{window function} that produces a smoothly varying response over the triangle plane. Its definition is as:
\begin{equation}
I(\mathbf{p}) = \mathrm{ReLU} \left( \frac{\phi(\mathbf{p})}{\phi(\mathbf{s})} \right)^{\sigma},
\end{equation}
where $\mathbf{s} \in \mathbb{R}^3$ is the incenter of triangle, $\sigma$ is the smoothness factor, and $\phi$ is defined as: 
\begin{equation}
\phi(\mathbf{p}) = \max_{i \in \{1,2,3\}} L_i(\mathbf{p}), \quad
L_i(\mathbf{p}) = \mathbf{n}_i \cdot \mathbf{p} + d_i,
\label{equation:windowfunction}
\end{equation}
where $\mathbf{n}_i$ is the unit normal of the $i^{th}$ edge pointing outward, and $d_i$ is the offset such that $L_i = 0$ when $\mathbf{p}$ lies on the $i^{th}$ edge.
\[
\quad \text{such that} \quad
I(\mathbf{p}) =
\begin{cases}
1 & \text{at the triangle incenter,} \\
0 & \text{at the triangle edge,} \\
0 & \text{outside the triangle.}
\end{cases}
\]
The smoothness factor $\sigma$ controls the sensitivity of the window function’s response. As $\sigma$ approaches zero, the triangle becomes nearly solid, and its response remains uniform for all points within the triangle. Conversely, as $\sigma$ increases, the response becomes more sensitive to variations within the triangle. As illustrated in Figure~\ref{fig:windowfunction}, this setting enables the triangle to be differentiable.
Unlike 3DTS~\cite{3dts}, we define our window function in world space, and we demonstrate that this definition is compatible with 3DTS. Figure~\ref{fig:compatibility} illustrates the result of directly rendering the triangles optimized by 3DTS's rasterization-based method using our tracer. 

\subsection{GPU-Accelerated Ray Tracing}
For GPU-accelerated ray tracing, we adopt the OptiX~\cite{OptiX} framework. The construction of an OptiX program can be roughly divided into two steps: building the acceleration structure (BVH) and defining the ray tracing behavior through various built-in programs. For BVH construction, since triangles are used as primitives, only the vertex array and an index buffer are required as inputs. This eliminates the need to define custom primitives and their bounding boxes, as required in~\cite{3dgrt}. 

For the triangle rendering part, we follow the method of~\cite{3dgrt, Knoll2019}. In general, within the OptiX \textbf{Ray generation} program, a ray is first emitted, and the intersection information of the closest $k$ triangles along the ray direction is stored in a buffer. This is achieved by performing an insertion sort within the \textbf{Any-hit} program. For further details on the k-closest intersection algorithm, please refer to~\cite{3dgrt}. Then, the colors of these $k$ triangles are accumulated along with the ray transmittance. depending on whether the accumulated transmittance of the ray exceeds a threshold or all triangles have been traversed, the ray tracing process either terminates or continues to the next intersection. This procedure is repeated iteratively until the stopping condition is met.

Different from rasterization, we do not project primitives into the image space using the projection matrix of a pinhole camera model. Instead, our tracer takes an externally provided array of ray origins and ray directions as input , which allows complex camera and lens models to be easily modeled.

\subsection{Triangle Optimization}
\begin{figure}
    \centering
    \includegraphics[width=\linewidth]{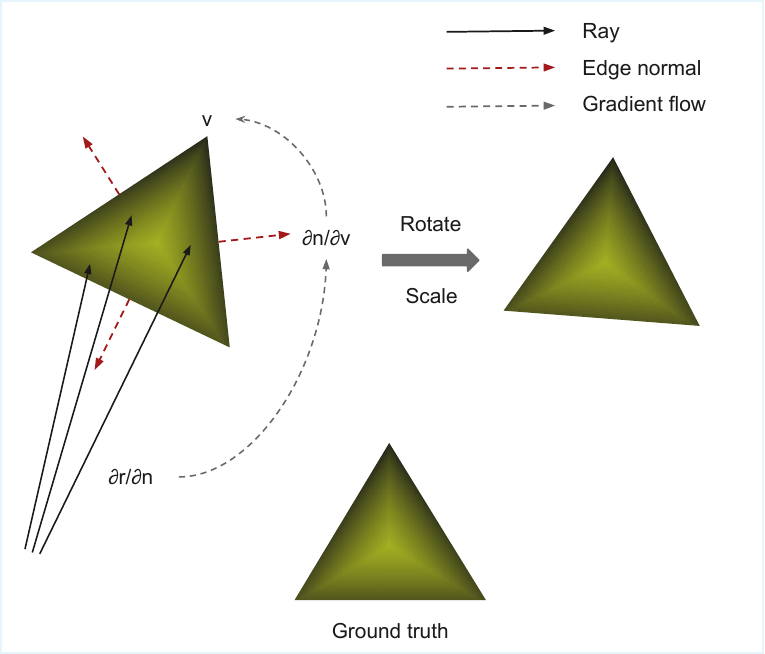}
    \caption{{\bf Gradient propagation on triangle vertices.} Gradients from the ray direction ($\partial r / \partial n$) propagate through the \textbf{window function} to the edge normals ($\partial n / \partial v$), updating vertex positions. This process results in triangle rotation and scaling toward the ground-truth shape. The detailed gradient derivations are provided in the \textit{supplementary material}.
}
    \label{fig:gradients}
\end{figure}

{\bf Gradients to triangle.} The optimizable parameters of a triangle include the spherical harmonics coefficients used to reconstruct the color $c$, the opacity, the smoothness factor $\sigma$, and the three vertices $\mathbf{v}_1$, $\mathbf{v}_2$, and $\mathbf{v}_3$. 
Unlike Gaussians or 2DGS~\cite{2dgs}, triangles do not have explicit position or scale parameters; therefore, all geometric properties are affected solely by changes in the spatial positions of the vertices. 
As a result, the design of the gradient propagation chain for the triangle vertices is crucial. 
After several attempts, we found that the following formulation yields stable and effective results: 
\begin{equation}
    \mathbf{N}_i=
    [(\mathbf{v}_i - \mathbf{v}_{i+2})\times(\mathbf{v}_{i+1}-\mathbf{v}_{i+2})]\times(\mathbf{v}_{i+1-}\mathbf{v}_i), 
    \label{equation:edgenormal}
\end{equation}
The unit edge normal in Equation~\ref{equation:windowfunction} is defined as $\mathbf{n}_i = \frac{\mathbf{N}_i}{\|\mathbf{N}_i\|}$. 
As described earlier, based on the definition of the window function in Equation~\ref{equation:windowfunction}, when $\sigma > 0$, the triangle has different responses to different points $\mathbf{p}$ inside it. 
According to the definition in Equation~\ref{equation:edgenormal}, these responses allow gradients to propagate from the window function to the vertices, enabling the optimizer to rotate and scale the triangle by adjusting the relative positions of its vertices to minimize the loss. 
Figure~\ref{fig:gradients} illustrates this process.

\noindent{\bf Pruning and densification.} We adopt the pruning and Densification strategy proposed by ~\cite{3dts}.
In this strategy, three factors determine which triangles are removed. 
The first factor is the opacity, a widely adopted criterion. 
Triangles with very low opacity contribute negligibly to the final pixel values and are therefore pruned. 

The second factor considers the product $ \omega = T \cdot o \cdot \rho$, where $T$ is the ray transmittance, $o$ is opacity and $\rho$ is the response of the window function. 
Triangles with $\omega < \omega_{\text{min}}$, where $\omega_{\text{min}}$ is a user-defined threshold, are pruned as well.
\begin{figure}[b]
    \centering
    \includegraphics[width=0.9\linewidth]{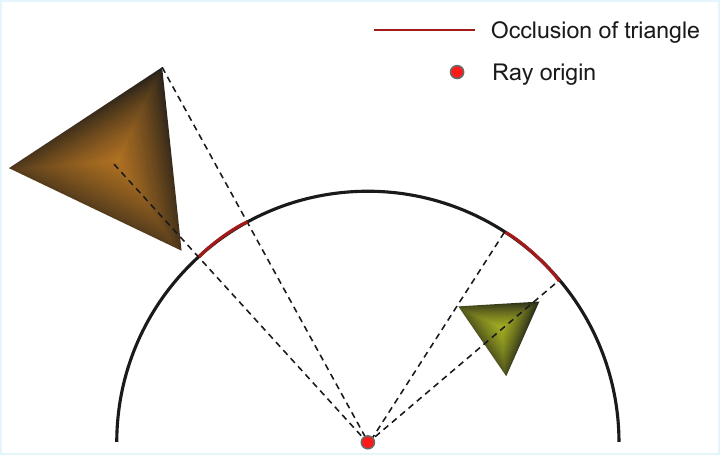}\\
    \caption{{\bf Occlusion comparison.} The distance of a triangle from the ray origin affects its occlusion. Although the triangles on the left and right differ significantly in absolute size, they exhibit similar levels of occlusion. }
    \label{fig:scaling}
\end{figure}
The third factor is the number of times a triangle is observed by different camera views. 
We modified this criterion such that, instead of measuring the triangles’ extent in the image space, we record whether a triangle is hit by any ray during ray tracing for a single camera view. If a triangle is hit by any number of rays within that view, its hit count is incremented by one. After all camera views have been processed, triangles with hit counts less than two are pruned.

We also modify how the triangle footprint is measured to cooperate with MCMC densification~\cite{3dgsmcmc}. In our densification stage, triangles that become too large are subdivided into four smaller triangles. 
3DTS measures footprint size in image space, computed from the distances between each vertex and the triangle’s centroid.
\begin{table*}[t]
\centering
\caption{{\bf Quantitative results.} Comparison of ours, baseline, and reference results on Mip-NeRF360~\cite{mignerf360} and Tanks \& Temples~\cite{tnt}. $\dagger$ denotes reproduced results, other results are reported in~\cite{3dts}.}
\label{tab:benchmark}
\resizebox{\textwidth}{!}{%
\begin{tabular}{l|ccc|ccc|ccc|ccc}
\toprule
& \multicolumn{3}{c|}{Mip-NeRF360 outdoor}
& \multicolumn{3}{c|}{Mip-NeRF360 indoor}
& \multicolumn{3}{c|}{Mip-NeRF360 average}
& \multicolumn{3}{c}{Tank \& Temples} \\
\cmidrule(lr){2-4}
\cmidrule(lr){5-7}
\cmidrule(lr){8-10}
\cmidrule(lr){11-13}
& PSNR $\uparrow$ & SSIM $\uparrow$ & LPIPS $\downarrow$
& PSNR $\uparrow$ & SSIM $\uparrow$ & LPIPS $\downarrow$
& PSNR $\uparrow$ & SSIM $\uparrow$ & LPIPS $\downarrow$
& PSNR $\uparrow$ & SSIM $\uparrow$ & LPIPS $\downarrow$ \\
\midrule
3DGS  & 26.40 & 0.805 & 0.173  & 30.41 & 0.920 & 0.192 & 28.69 & 0.870 & 0.182 & 23.14 & 0.841 & 0.183 \\
2DGS  & 26.10 & 0.790 & 0.185  & 30.41 & 0.916 & 0.195 & 28.56 & 0.862 & 0.190 & 23.13 & 0.832 & 0.212 \\
3DTS$\dagger$   & 26.34 & 0.800 & 0.170  & 30.91 & 0.933 & 0.141 & 28.95 & 0.876 & 0.153 & 23.06 & 0.842 & 0.164 \\
\midrule
3DGRT$\dagger$  & 25.92          & \textbf{0.784} & 0.221          & 30.12          & 0.916          & 0.245         & 28.32          & 0.859          & 0.235          & 22.76          & 0.844          & 0.201 \\
Ours            & \textbf{25.94} & 0.779          & \textbf{0.186} & \textbf{30.77} & \textbf{0.931} & \textbf{0.146} & \textbf{28.70} & \textbf{0.866} & \textbf{0.163} & \textbf{22.88} & \textbf{0.849} & \textbf{0.150} \\
\bottomrule
\end{tabular}
}
\end{table*}
However, since our optimization operates in world space rather than image space, such a measure is not directly applicable. Therefore, we introduce a world-space metric that reflects the triangle’s occlusion with respect to the camera. Specifically, we measure the angle between the vector from each vertex to the ray origin and the vector from the triangle’s centroid to the ray origin. This metric inherently accounts for distance: several small triangles that are very close to the camera will be regarded as large triangles in terms of occlusion, while distant triangles will have the opposite effect. 
Figure~\ref{fig:scaling} illustrates how this metric compares the occlusion effects of triangles at different distances.

\subsection{Training} 
Our training pipeline is implemented based on PyTorch~\cite{torch}, 
while the gradient computation of the loss with respect to the optimizable parameters is handled by custom CUDA kernels. 
The core component of our system is a differentiable ray tracer implemented with OptiX \cite{OptiX}. 
Following the implementations of \cite{Kerbl2023-nj,3dts,2dgs}, our overall loss function is defined as:
\begin{equation}
\mathcal{L} = (1 - \lambda_{c})\mathcal{L}_{1} 
+ \lambda_{c} \mathcal{L}_{\text{D-SSIM}} 
+ \lambda_{o}\mathcal{L}_{o} 
+ \lambda_{n}\mathcal{L}_{n} 
+ \lambda_{s}\mathcal{L}_{s}.
\end{equation}
We combine the $\mathcal{L}_1$ loss with the D-SSIM term \cite{Kerbl2023-nj}, 
the normal loss $\mathcal{L}_n$ \cite{2dgs}, 
and the opacity loss $\mathcal{L}_o$ and size loss $\mathcal{L}_s$ from\cite{3dts}.
$\mathcal{L}_s$ encourages larger triangles and is defined as:
$
 \mathcal{L}_s = 2 \cdot \left\| (\mathbf{v}_1 - \mathbf{v}_0) \times (\mathbf{v}_2 - \mathbf{v}_0) \right\|_{2}^{-1}.
$


\section{Experiments}
\label{sec:experiments}
\subsection{Datasets and Metrics}
\begin{figure*}
    \centering
    \includegraphics[width=\textwidth,trim=40 215 40 20,clip]{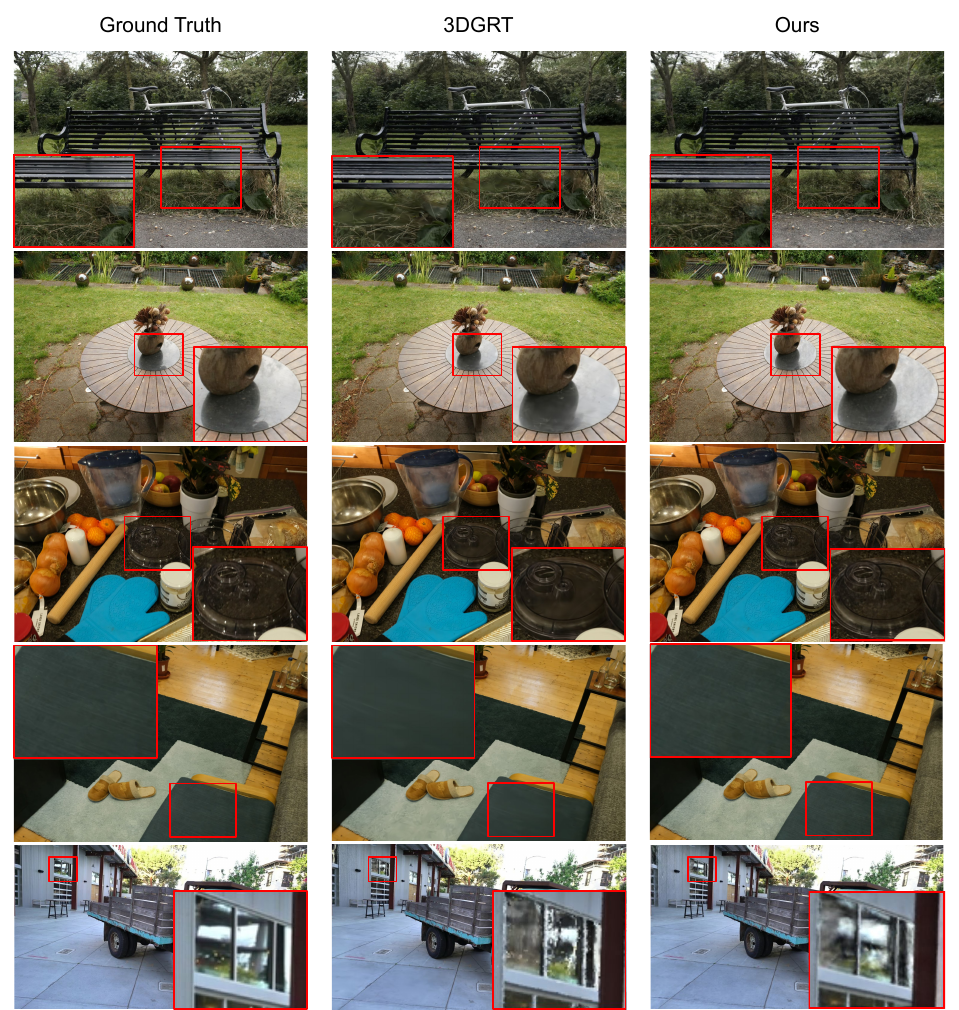}\\
    \caption{\textbf{Novel-View Synthesis Quality Comparison.} In the novel-view synthesis results, 3DGRT produces overly smooth appearances in high-frequency regions, leading to the loss of fine details. In the \textit{truck} scene, it introduces high-frequency noise in distant areas. Our method does not exhibit these issues.
}
    \label{fig:novelview}
\end{figure*}
For evaluation, we use the Mip-NeRF 360~\cite{mignerf360} and Tanks and Temples~\cite{tnt} datasets. 
To avoid license issues, we only use the following scenes from Mip-NeRF 360:
\textbf{outdoor}: \textit{bicycle}, \textit{garden}, \textit{stump};
\textbf{indoor}: \textit{room}, \textit{counter}, \textit{kitchen}, \textit{bonsai}. 
For these scenes, we use images down-sampled by a factor of two (indoor) and a factor of four (outdoor). In the Tanks and Temples dataset, we use the \textit{Train} and \textit{Truck} scenes.

We consider 3DGRT~\cite{3dgrt} to be the most closely related method, and we also include 3DGS~\cite{Kerbl2023-nj}, 2DGS~\cite{2dgs}, and Triangle Splatting for comparison. 
For quantitative evaluation, we report PSNR, SSIM~\cite{ssim}, and LPIPS~\cite{lpips}. 
In addition, we compare the average rendering speed on a single NVIDIA RTX 6000 Ada GPU. The implementation details and hyperparameters are provided in the \textit{supplementary material}.

\subsection{Results and Comparison}
We present quantitative results in Table~\ref{tab:benchmark} and qualitative comparisons in Figure~\ref{fig:novelview}.
Across all test scenes, our method consistently achieves better results than 3DGRT. 
While the PSNR values of the two methods are very close in outdoor scenes, it is worth noting that 3DGRT uses smooth Gaussian kernels as primitives, which tend to produce overly smoothed regions in novel views.
This smoothing effect can artificially increase PSNR.
In contrast, our method clearly outperforms 3DGRT in LPIPS, achieving an improvement of approximately 15\%, which indicates substantially better perceptual quality and detail preservation. 
We also include rasterization-based methods for comparison. 
Among them, 3DGS achieves higher PSNR than planar-primitive approaches such as 3DTS and 2DGS, which aligns with expectations.

For visualization, we select two outdoor scenes (\textit{bicycle}, \textit{garden}), two indoor scenes (\textit{counter}, \textit{room}) from the Mip-NeRF 360 dataset , and a large outdoor scene (\textit{truck}) from Tanks and Temples.
In these scenes, 3DGRT struggles to reproduce certain fine textures and even introduces high-frequency color noise (e.g., the glass region highlighted in red in \textit{truck}~\ref{fig:novelview}), whereas our method does not exhibit such artifacts.

\begin{figure}
    \centering
    \includegraphics[width=\linewidth,trim=120 40 120 120,clip]{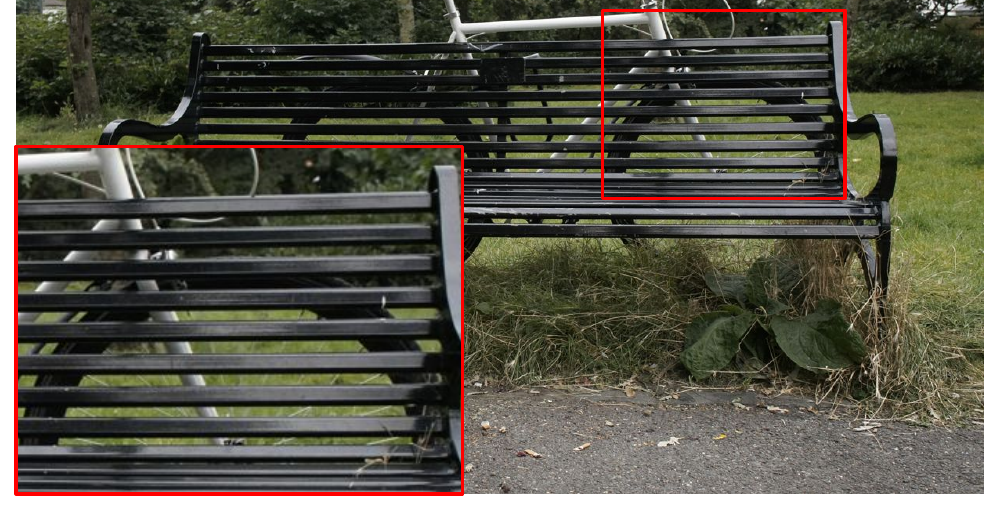}\\
    \includegraphics[width=\linewidth,trim=120 40 120 120,clip]{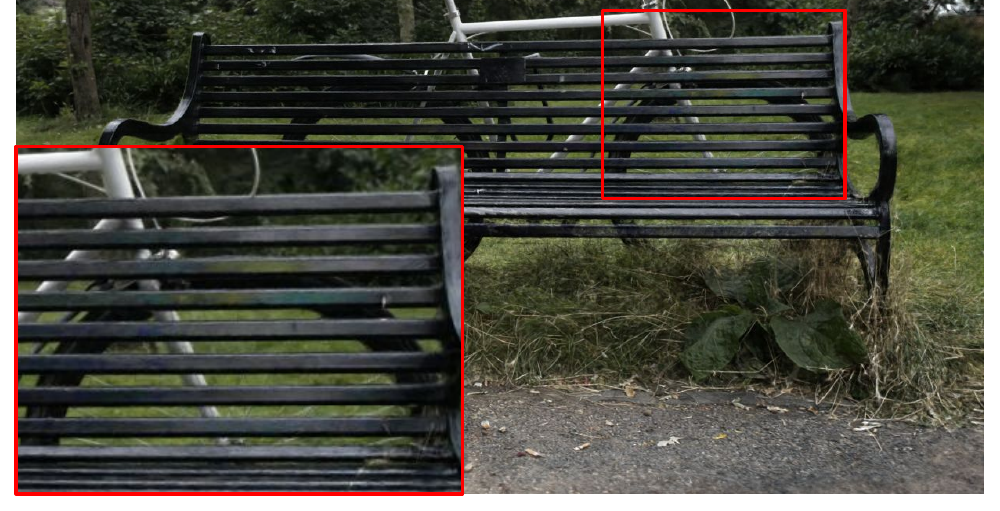}\\
    \includegraphics[width=\linewidth,trim=120 40 120 120,clip]{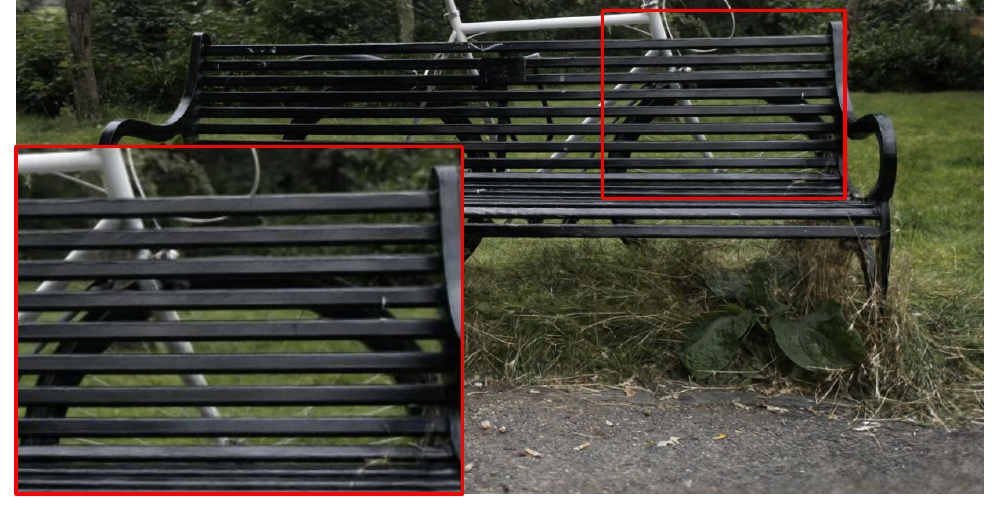}\\
    \caption{\textbf{Novel-view color comparison.} Top: Ground truth. Middle: 3DTS\cite{3dts}. Bottom: Ours. The method in 3DTS produces noisy colors, while ours maintains consistent colors.
}
\vspace{-1em}
    \label{fig:raydirection}
\end{figure}

We also observe that our method achieves more accurate color reproduction under challenging lighting conditions, compared to rasterization-based methods. 
For example, Triangle Splatting~\cite{3dts} produces less accurate results in novel-view synthesis, as shown in Figure~\ref{fig:raydirection}.
\begin{table}[b]
\centering
\caption{{\bf Rendering speed.} Comparison with baseline methods under two settings. $\dagger$ denotes results reported in~\cite{3dgrt}.}
\label{tab:renderperf}
\resizebox{\linewidth}{!}{%
\begin{tabular}{l|cc}
\toprule
& \multicolumn{2}{c}{FPS$\uparrow$} \\
\cmidrule(lr){2-3}
& Mip-NeRF360 & Tanks \& Temples \\
\midrule
3DGRT (performance)$\dagger$   & 78  & 190  \\
3DGRT (quality)$\dagger$     & 55  & 143  \\
\midrule
Ours                 & 37  & 119 \\
\bottomrule
\end{tabular}
}
\vspace{-1em}
\end{table}
\subsection{Ray Tracing Performance}
We report the ray tracing performance results in Table~\ref{tab:renderperf}. 
For 3DGRT, we evaluate two settings: \textit{quality} and \textit{performance}. 
The \textit{quality} setting uses the regular 3D Gaussian kernel and keeps the same optimization hyperparameters as in \cite{Kerbl2023-nj}. 
The \textit{performance} setting instead uses a degree-2 generalized Gaussian kernel~\cite{3dgrt} and adjusts the optimization hyperparameters to trade rendering quality for faster ray tracing. 
Our tracer, on the other hand, is evaluated using the quality-oriented configuration, including increasing the maximum number of allowed primitives. 
Despite this, our rendering speed is only about 30\% slower than 3DGRT (quality), which we consider acceptable for near real-time usage. It is worth noting that our ray tracing pipeline is currently unoptimized; with further engineering optimizations, we expect the rendering performance to improve significantly.

\subsection{Ablation study}
Table~\ref{tab:ablation} shows the ablation study results on the Mip-NeRF 360 dataset. 

\noindent\textbf{Loss term.} 
In our ray tracing implementation, we use the gradients from the regularization term $\mathcal{L}_n$ to optimize the triangle vertices. 
To evaluate the effect of this term, we remove $\mathcal{L}_n$ during training. 
We observe that under our current setting of $\lambda_n$, removing this loss term leads to a slight degradation in the final rendering quality.

In addition, we specifically remove $\mathcal{L}_s$ to examine its impact on the rendering results. 
We observe that removing it leads to a degradation in overall rendering quality and about a 0.1\% increase in the number of triangles. 
This is because more small triangles are required to fill the empty regions that would otherwise be covered by a few large triangles.

\noindent\textbf{World space occlusion.}  
We remove our modification to the triangle footprint metric and instead use a constant footprint value of zero. Without the world space occlusion metric, MCMC densification~\cite{3dgsmcmc} is unable to distinguish between small and large triangles. In the first scene (\textit{bicycle}), the training speed drops by more than \(5\times\), and the optimization fails to converge. We therefore consider this configuration as unable to complete training.

\noindent\textbf{View pruning.}  
We disable the pruning of triangles that intersect rays from only a single view. 
As a consequence, nearly degenerate triangles are still included in gradient computation. 
Due to multiple multiplication operations in our vertex gradient propagation, extremely small gradients may underflow and produce NaN values. 
In the \textit{stump} scene, NaN values appear on triangle vertices, causing the training to terminate.

\begin{table}
\centering
\caption{{\bf Ablation Study.} We vary one component at a time to evaluate its impact on rendering quality.}
\label{tab:ablation}
\resizebox{\linewidth}{!}{%
\begin{tabular}{l|ccc}
\toprule
& PSNR $\uparrow$ & SSIM $\uparrow$ & LPIPS $\downarrow$ \\
\midrule
Ours                      & 28.70 & 0.866 & 0.163  \\
\midrule
w/o world space occlusion & \multicolumn{3}{c}{N/A} \\
w/o view pruning          & \multicolumn{3}{c}{N/A} \\
w/o $\mathcal{L}_n$       & 28.69 & 0.865 & 0.163 \\
w/o $\mathcal{L}_s$       & 28.54 & 0.864 & 0.164 \\
\bottomrule
\end{tabular}
}
\end{table}

\section{Conclusion}
In this paper, we introduce \textbf{UTrice}, a novel differentiable ray tracing framework that uses triangles as primitives for particle-based 3D scenes.
By directly tracing triangles instead of relying on proxy geometry, our method removes redundant BVH construction and custom intersection logic, eliminating significant overhead and enabling higher rendering quality. It can also directly render triangles optimized via rasterization without any preprocessing, allowing fast rasterization-based optimization to be combined with ray tracing effects. In this way, our approach unifies rasterization and ray tracing within a single pipeline for particle-based scene optimization.
Moreover, since our ray tracer only requires rays as input, it is conceptually compatible with non-pinhole camera models, such as LiDAR sensors.

\noindent{\bf Limitation.} While our approach achieves higher rendering quality, it produces a larger number of primitives. Because the generated triangle soup has no mesh connectivity, neighboring vertices are redundantly stored, increasing memory usage and computational load. The training pipeline is also unoptimized and contains computational redundancy, making training roughly \(2\times\) slower than 3DGRT~\cite{3dgrt}; moreover, it currently lacks a mechanism for handling extreme triangles. We leave reducing primitive count, eliminating redundant storage, and improving robustness to extreme triangles as future work

\iftoggle{cvprfinal}{
\section*{Acknowledgments}
We would like to express our gratitude to Professor Nobuyuki Umetani for his invaluable guidance throughout this work.

}

{
    \small
    \bibliographystyle{ieeenat_fullname}
    \bibliography{main}
}

\clearpage

\setcounter{page}{1}
\maketitlesupplementary
\appendix 
\section{Gradient Computations}
\begin{figure}[b]
    \centering
    \includegraphics[width=.8\linewidth,trim=160 10 160 40,clip]{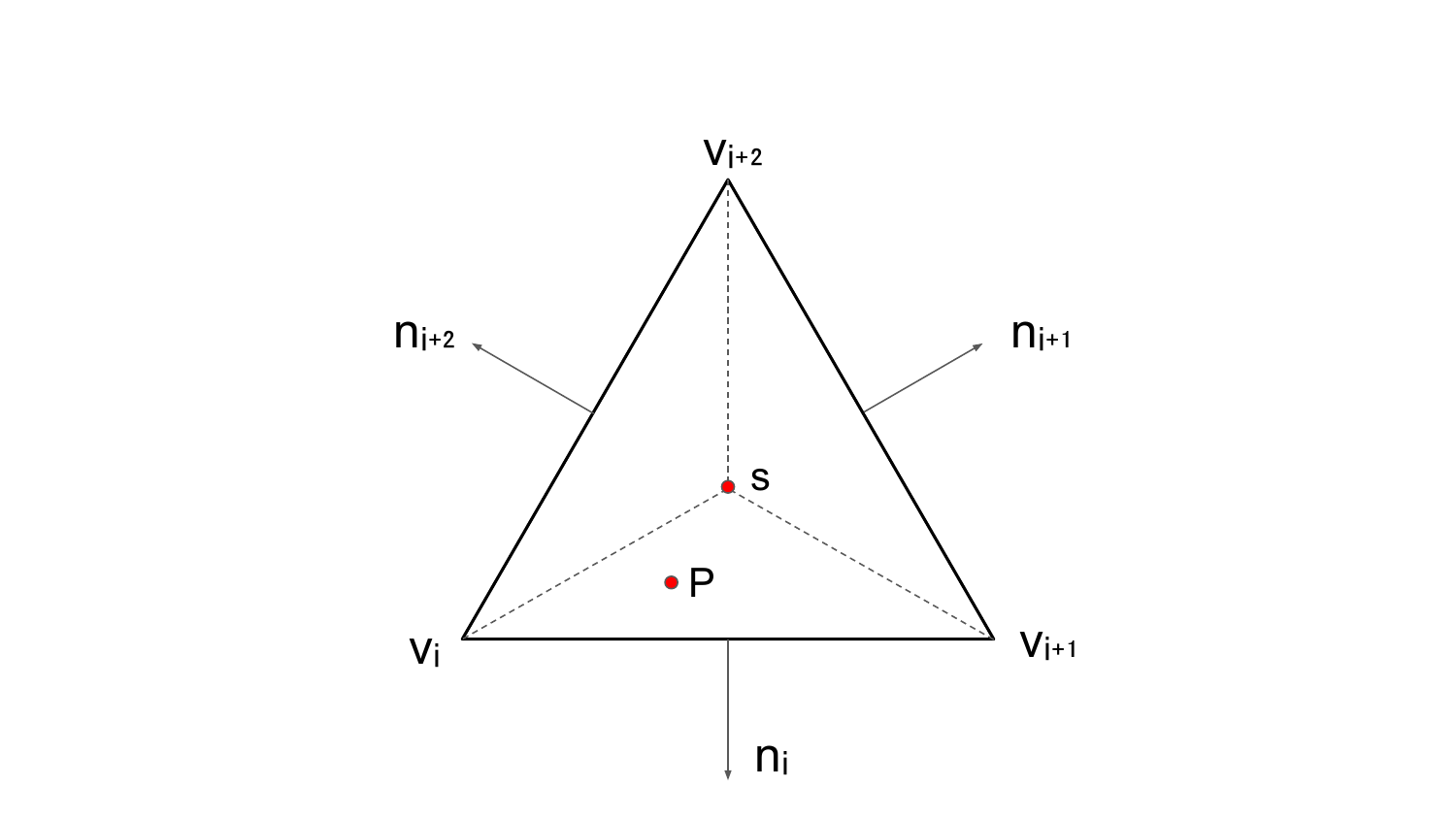}
    \caption{ {\textbf{Geometric configuration of the triangle.}}}
    \label{fig:triangle}
\end{figure}
For the triangle shown in Figure~\ref{fig:triangle}, based on the definition of the window function in Section 3.3, we define
\[
a = \mathbf{v}_i - \mathbf{v}_{i+2}, \
b = \mathbf{v}_{i+1} - \mathbf{v}_{i+2}, \
c = \mathbf{v}_i - \mathbf{v}_{i+1}, \
d_i = -\mathbf{n}_i \cdot \mathbf{v}_i.
\]
Here,
\[
i = \argmax_{i\in\{1,2,3\}} L_i(\mathbf{p}), \
\phi(\mathbf{p}) = L_{i}(\mathbf{p}).
\]
With the above definitions, the gradients of the edge normal $\mathbf{n}_i$ w.r.t. the triangle vertices are given by
\[
\begin{aligned}
\frac{d\mathbf{n}_i}{d\mathbf{v}_i} &= b c^\top - (b\cdot c) \mathbf{I} + [a\times b]_\times \\
\frac{d\mathbf{n}_i}{d\mathbf{v}_{i+1}} &= -\!\left( a c^\top - (a\cdot c)\mathbf{I} + [a\times b]_\times \right) \\
\frac{d\mathbf{n}_i}{d\mathbf{v}_{i+2}} &= c c^\top - (c\cdot c)\mathbf{I}
\end{aligned}
\]
where $[a\times b]_\times$ denotes the skew\text{-}symmetric matrix of $a\times b$.

\noindent Finally, the gradients of $\mathcal{L}$ w.r.t. vertices are
\[
\begin{aligned}
\frac{d\mathcal{L}}{d\mathbf{v}_i}
&=
\frac{d\mathcal{L}}{dI}
\left(
\frac{dI}{d\phi(\mathbf{p})} 
\frac{d\phi(\mathbf{p})}{d\mathbf{v}_i}
+
\frac{dI}{d\phi(\mathbf{s})} 
\frac{d\phi(\mathbf{s})}{d\mathbf{v}_i} 
\right) \\
&\approx 
\frac{d\mathcal{L}}{dI}
\frac{dI}{d\phi(\mathbf{p})}
\frac{d\phi(\mathbf{p})}{d\mathbf{v}_i} \\
&=
\frac{d\mathcal{L}}{d\phi(\mathbf{p})}
\left(
(\mathbf{p}-\mathbf{v}_i)\frac{d\mathbf{n}_i}{d\mathbf{v}_i} - \mathbf{n}_i
\right) \\
\frac{d\mathcal{L}}{d\mathbf{v}_{i+1}}
&\approx
\frac{d\mathcal{L}}{d\phi(\mathbf{p})}
(\mathbf{p}-\mathbf{v}_{i})\frac{d\mathbf{n}_i}{d\mathbf{v}_{i+1}} \\
\frac{d\mathcal{L}}{d\mathbf{v}_{i+2}}
&\approx
\frac{d\mathcal{L}}{d\phi(\mathbf{p})}
(\mathbf{p}-\mathbf{v}_{i})\frac{d\mathbf{n}_i}{d\mathbf{v}_{i+2}} \\
\end{aligned}
\]
\begin{figure*}
    \centering
    \includegraphics[width=\linewidth,trim=5 300 8 0,clip]{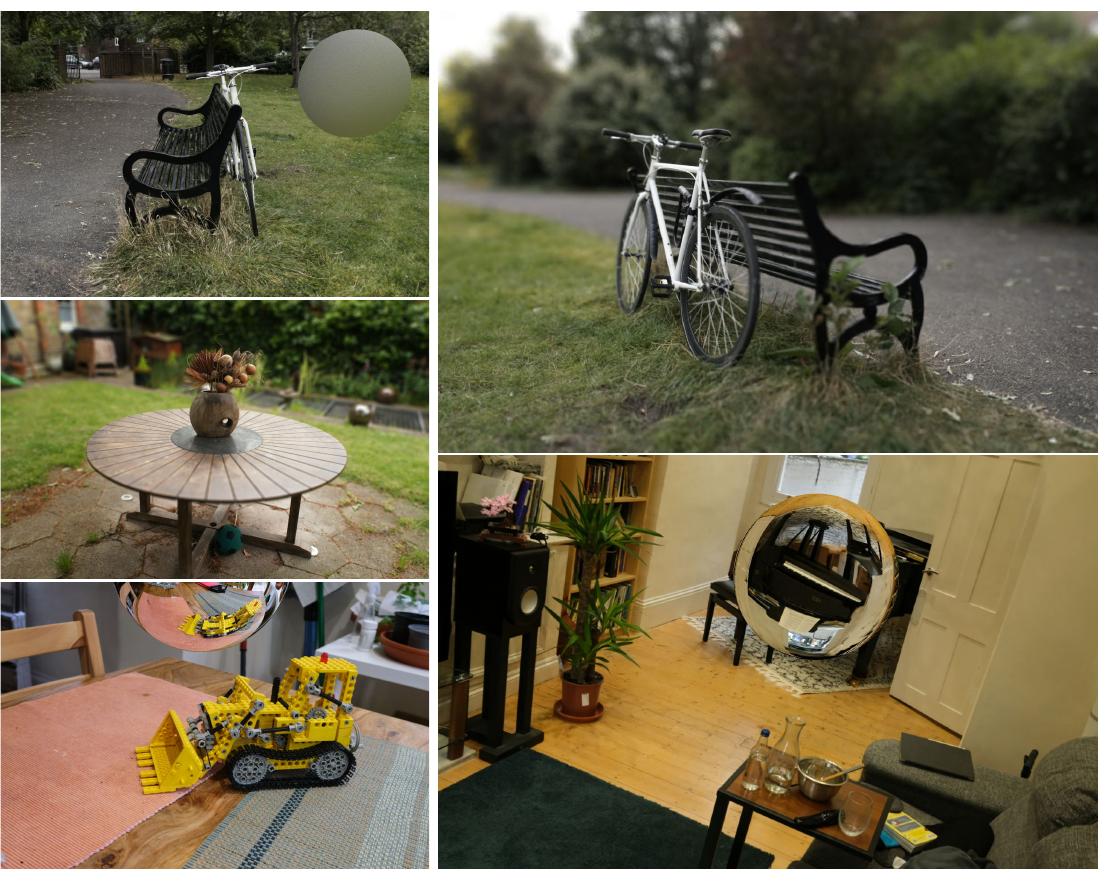}
        \vspace{-2em}
        \caption{{\bf Ray tracing effects.} Once the scene is optimized, various ray tracing effects can be easily achieved. We illustrate several examples, including environment lighting, depth of field, reflection, and refraction.}
    \label{fig:effects}
\end{figure*}
\section{Hyperparameters and Training Details}
\begin{table}[b]
\centering
\caption{\bf Hyperparameters of Ours and 3DTS.}
\label{tab:hyperparams_our}
\resizebox{0.9\linewidth}{!}{%
\begin{tabular}{l|cc}
\toprule
& Ours & 3DTS \\
\midrule
feature\_lr & 0.0025 & 0.0025 \\
opacity\_lr & 0.014 & 0.014 \\
lambda\_normals  & 0.0001 & 0.0001 \\
lambda\_opacity & 0.0055 & 0.0055 \\
lambda\_size &  1e-8 & 1e-8 \\
opacity\_dead & 0.014 & 0.014  \\
importance\_threshold & 0.022 & 0.022 \\
lr\_sigma & 0.0008 & 0.0008 \\
lr\_triangles\_points\_init & 0.0011 & 0.001 \\
split\_size & 0.019 & 24.0 \\
max\_noise\_factor & 1.5 & 1.5 \\
densification\_interval & 500  & 500 \\
densify\_from\_iter & 500 & 500 \\
densify\_until\_iter & 25000 & 25000 \\
add\_shape & 1.3 & 1.3 \\
\bottomrule
\end{tabular}
}
\end{table}
\begin{table}
\centering
\caption{\bf Hyperparameters of 3DGRT.}
\label{tab:hyperparams_3dgrt}
\resizebox{0.7\linewidth}{!}{%
\begin{tabular}{l|c}
\toprule
& 3DGRT \\
\midrule
features\_albedo\_lr & 0.0025 \\
features\_specular\_lr & 0.05 \\
density\_lr & 0.05 \\
rotation\_lr & 0.001 \\
scale\_lr & 0.005 \\
densify frequency & 300 \\
densify start\_iteration & 500 \\
densify end\_iteration & 15000 \\
prune frequency & 100 \\
prune start\_iteration & 500 \\
prune end\_iteration & 15000 \\
\bottomrule
\end{tabular}
}
\end{table}
\begin{table}
\centering
\caption{\bf Number of primitives (in millions) at the 30k iteration checkpoint.}
\label{tab:size}
\resizebox{\linewidth}{!}{%
\begin{tabular}{l|c|cc}
\toprule
& \multicolumn{3}{c}{\#Primitives $\downarrow$} \\
\cmidrule(lr){2-4}
& 3DTS & 3DGRT & Ours  \\
\midrule
bicycle & 4.75 & 6.33  & 5.14 \\
garden & 4.47  & 4.11 & 4.84 \\
stump & 4.25 & 7.64  & 4.48 \\
\midrule
bonsai  & 2.60 & 1.50 & 2.27 \\
counter & 2.33  & 1.26 & 2.37 \\
room  & 1.89  & 1.18 & 1.94 \\
kitchen  & 2.23  & 1.51 & 2.23 \\
\midrule
Mip-NeRF360 average & 3.22 & 3.36  & \bf{3.32} \\
\midrule
train & 2.35 & 3.53  & 2.43 \\
truck & 1.86 & 4.22  & 1.95 \\
\midrule
Tank \& Temples average  & 2.10  & 3.88& \bf{2.19} \\
\bottomrule
\end{tabular}
}
\end{table}
The hyperparameters used in our experiments are shown in
Table~\ref{tab:hyperparams_our}. We use the same settings for the
\textit{indoor} and \textit{outdoor} scenes of Mip-NeRF360~\cite{mipnerf} as well as for the
Tanks \& Temples dataset~\cite{tnt}. Table~\ref{tab:hyperparams_3dgrt}
lists the hyperparameters used for 3DGRT.

For our method and for 3DTS, we set an upper bound on the number of
triangles for each scene, and this bound is kept identical for both
methods. For 3DGRT, we do not impose such a limit and simply reuse all hyperparameters specified in the original \textit{base\_gs.yaml} configuration.

In Table~\ref{tab:size}, we report the number of primitives contained in each method at the 30k iteration checkpoint.

All 3DGRT results in our experiments are reproduced using the example code provided in the official implementation~\cite{3dgrt}, and the 3DTS results are similarly obtained using the example code from~\cite{3dts}.
We directly record the metric values reported by each official implementation.
\section{Additional Rendering Results}
In our implementation, common ray tracing effects can be easily integrated, as shown in Figure~\ref{fig:effects}.
We apply a single bounce for both reflected and refracted rays rather than a full dielectric BSDF model. For depth of field, we employ distributed ray tracing with a variable number of samples.
%
%
%
%

\end{document}